\documentclass[10pt,twocolumn,letterpaper]{article}

\usepackage{cvpr}
\usepackage{times}
\usepackage{epsfig}
\usepackage{graphicx}
\usepackage{amsmath}
\usepackage{amssymb}

\usepackage{multirow}
\usepackage{array}
\usepackage{makecell}
\usepackage{booktabs}
\usepackage{gensymb}
\usepackage[ruled]{algorithm2e}

\newcommand{\PreserveBackslash}[1]{\let\temp=\\#1\let\\=\temp}
\newcolumntype{C}[1]{>{\PreserveBackslash\centering}p{#1}}
\newcolumntype{R}[1]{>{\PreserveBackslash\raggedleft}p{#1}}
\newcolumntype{L}[1]{>{\PreserveBackslash\raggedright}p{#1}}
\usepackage[breaklinks=true,bookmarks=false]{hyperref}

\cvprfinalcopy 

\def\cvprPaperID{2365} 

\ifcvprfinal\fi
\begin{document}
\title{PFRL: Pose-Free Reinforcement Learning for 6D Pose Estimation}
\author{
\begin{tabular}{ccccc}
Jianzhun Shao &
Yuhang Jiang &
Gu Wang &
Zhigang Li &
Xiangyang Ji\\
\multicolumn{5}{c}{Department of Automation and BNRist, Tsinghua University}\\
\multicolumn{5}{c}{Beijing, China}\\
\multicolumn{5}{c}{\small \{sjz18,jiangyh19,wangg16,lzg15\}@mails.tsinghua.edu.cn \small \qquad xyji@tsinghua.edu.cn}\\ 
\end{tabular}\\[3ex]
}

\maketitle
\begin{abstract}
   6D pose estimation from a single RGB image is a challenging and vital task in computer vision. The current mainstream deep model methods resort to 2D images annotated with real-world ground-truth 6D object poses, whose collection is fairly cumbersome and expensive, even unavailable in many cases. In this work, to get rid of the burden of 6D annotations, we formulate the 6D pose refinement as a Markov Decision Process and impose on the reinforcement learning approach with only 2D image annotations as weakly-supervised 6D pose information, via a delicate reward definition and a composite reinforced optimization method for efficient and effective policy training. Experiments on LINEMOD and T-LESS datasets demonstrate that our Pose-Free approach is able to achieve state-of-the-art performance compared with the methods without using real-world ground-truth 6D pose labels.
\end{abstract}
\section{Introduction}
6D pose estimation aims to localize the 3D location and the 3D orientation of the object from a single image. It plays a crucial role in real-world applications including robot manipulation~\cite{collet2011moped,zhu2014single}, augmented reality~\cite{marchand2015pose} and self-driving cars~\cite{chen2017multi,manhardt2019roi}. 
For instance, when a robot tries to grab an object, it is a prerequisite to accurately estimate the 6D object poses from the image captured by the equipped camera.

This problem was traditionally regarded as a geometric problem and can be solved with Perspective-n-Point (PnP) \cite{lepetit2009epnp} algorithms by matching features between the 2D image and the 3D object model. 
Since rich textures are indispensable for feature matching, they cannot effectively handle the texture-less cases. 
With the rise of deep learning, considerable learning-based approaches are proposed for various perception cases (e.g., RGB-only, RGB-D) and have greatly promoted the evolvement of this field. 
Some people \cite{pavlakos20176,rad2017bb8,peng2019pvnet} followed the conventional way to build the 2D-3D correspondences and subsequently solve the pose via PnP. 
Others \cite{kehl2017ssd,xiang2017posecnn} instead trained the deep model in an end-to-end manner to directly derive the pose from the image. 
Compared with traditional ones, these approaches can achieve remarkable performance even in challenging situations such as texture-less, clutter, occlusion, etc.  
A common way to further improve the pose accuracy is pose refinement. DeepIM \cite{li2018deepim} is able to predict the 6D pose difference between the current estimation and the ground-truth by comparing the rendered object with the observed one in 2D space. A similar idea is also explored by \cite{manhardt2018deep,zakharov2019dpod}. 

\begin{figure}[t]
\begin{center}
\includegraphics[height=4cm]{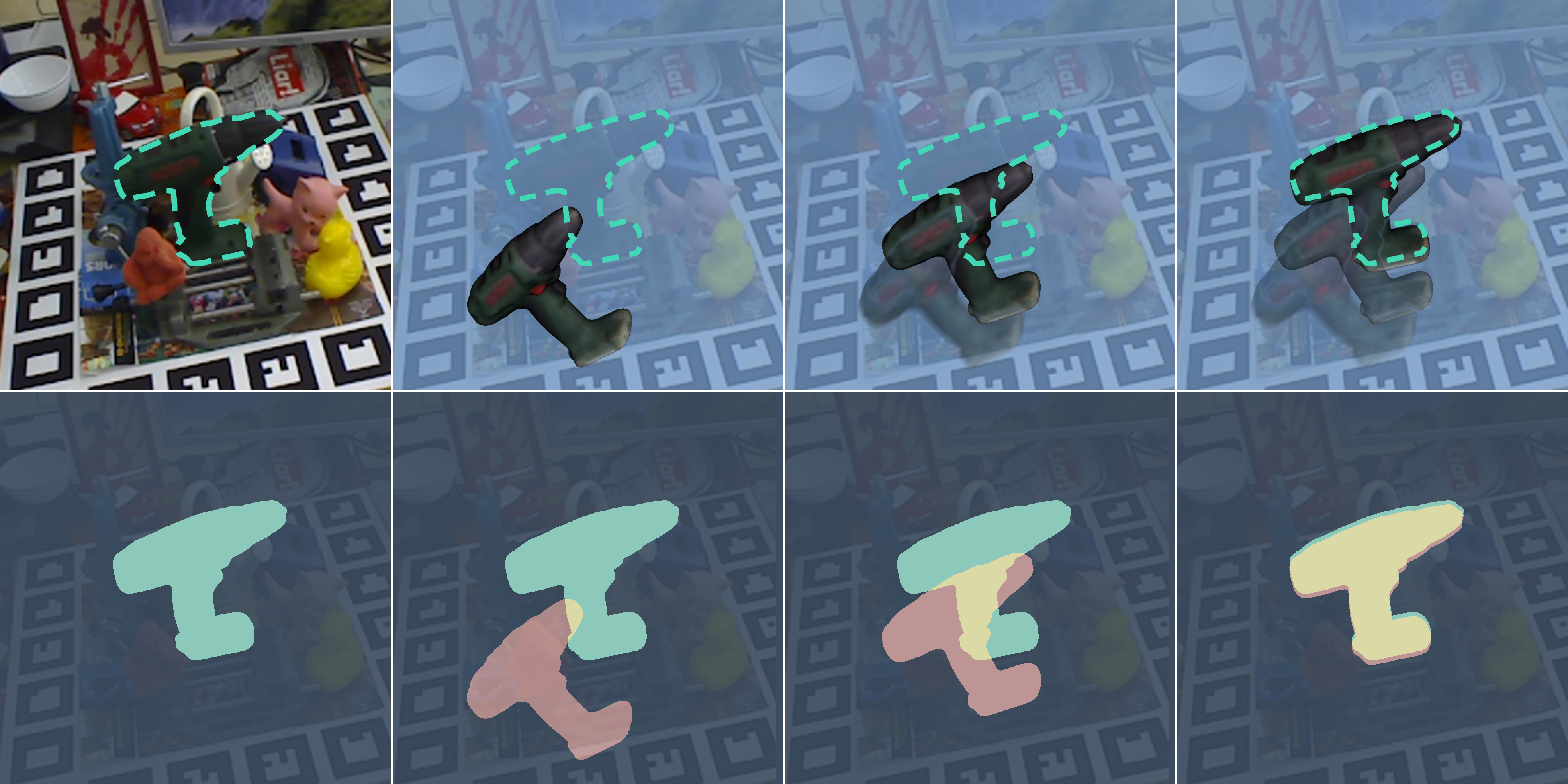}
\end{center}
\caption{The sketch of Pose-Free Reinforcement Learning (PFRL). We formulate the pose refining procedure as a sequential decision-making process and exploit a composite reinforced optimization approach to tackle it. We calculate the reward using the similarity measurement  of mask information to avoid the annotations of pose labels.}
\label{overall}
\end{figure}

Real-world pose annotations are essential for current approaches to achieve excellent performance. 
Unfortunately, acquisition of the 2D images annotated with real-world ground-truth 6D object poses is fairly cumbersome and expensive compared with that in 2D space (e.g. instance 2D mask), which greatly restricts the popularization of these approaches. 
Alternatively, some researchers \cite{kehl2017ssd,sundermeyer2018implicit,zakharov2019dpod} turned to getting rid of this problem by training on synthetic data only. 
However, the domain gap is prone to degrade pose estimation performance in real-world cases.

In this work, we focus on 6D pose estimation from RGB images only without requiring real-world pose annotations and avoiding the domain gap caused by training the model on synthetic-only data.

To achieve this goal, we introduce reinforcement learning by formulating the 6D pose refinement problem as a Markov Decision Process (MDP). Given an initial pose estimate, our approach learns to rotate and translate the 3D object model sequentially to align it to the observation. To train the model, we design a 2D mask-based reward which is computed only from the rendered and the target 2D masks. For each episode, the accumulated rewards are maximized, resulting in a policy that can approach the target smoothly and quickly (Fig. 
\ref{overall}). We also propose a composite reinforced optimization method to learn the operation policy efficiently and effectively. Compared with existing supervised approaches, our method utilizes a sequential decision-making process to optimize a delayed accumulated reward from the weakly-supervised mask similarity instead of the supervised loss from the ground-truth 6D pose. Since no ground-truth 6D pose information is involved, we call our proposed approach a Pose-Free Reinforcement Learning (PFRL) approach for 6D pose estimation.

Experiments on LINEMOD \cite{hinterstoisser2012model} and T-LESS \cite{hodan2017t} datasets demonstrate that our approach is able to achieve state-of-the-art performance compared with existing methods without using ground-truth 6D pose labels. 
Our work makes the following main contributions: 
i) We formulate the 6D pose estimation problem as a Markov Decision Process and propose a Pose-Free Reinforced Learning solution, which is able to exploit 2D image annotations as weakly supervised information and reward-based sequential decision making for 6D pose refinement.
ii) We design a low-cost reward strategy based on the 2D mask and propose a composite reinforced optimization method for efficient and effective policy training.
iii) On LINEMOD and T-LESS datasets we achieve state-of-the-art performance compared to the methods without using real-world ground-truth 6D pose labels.

\begin{figure*}[!htbp]
\begin{center}
\includegraphics[width=\textwidth]{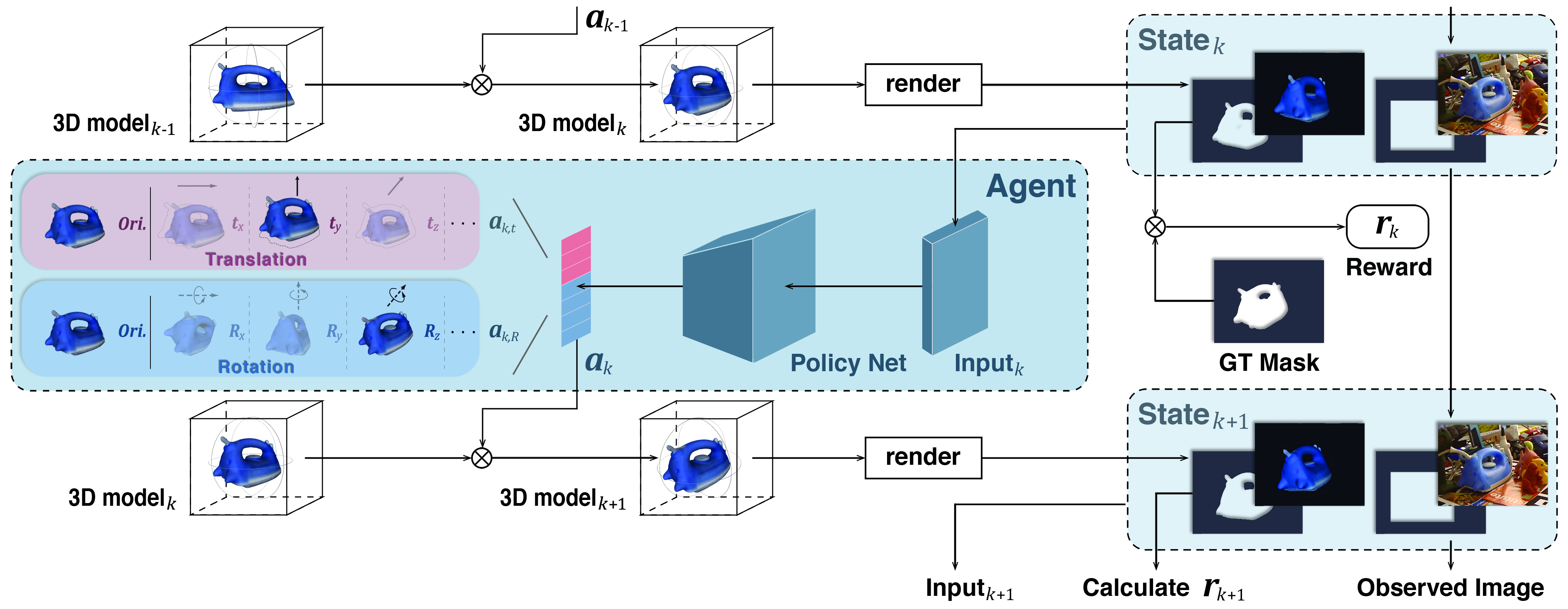}
\end{center}
\caption{The PFRL framework. At each time step $k$, we use the cropped observed image, ground-truth bounding box, rendered image and rendered mask  to form state $\mathbf{s}_k$. 
The policy network (detailed in Sec. \ref{netarc}) takes $\mathbf{s}_k$ as input and generates disentangled action $\mathbf{a}_k$, which represents the relative SE(3) transformation for current pose (detailed in Sec. \ref{rlsetting}). 
The environment (i.e. renderer) then renders a new image according to the new pose, calculating the reward from current mask and the ground-truth mask (detailed in Sec. \ref{reward}). 
Finally the composite policy optimization is executed using the reward (detailed in Sec. \ref{rlalgo}).}
\label{rlp}
\end{figure*}
\section{Related Work}
\textbf{RGB-based 6D Object Pose Estimation}.
Traditionally, the pose estimation was considered as a geometric problem and solved with Perspective-n-Point (PnP) \cite{lepetit2009epnp} algorithms by matching the features between the 2D image and the 3D model. 
However, they rely heavily on the textures and cannot handle the textures-less cases. 
Recently, the deep learning technique has significantly advanced the development of pose estimation. 
Some works followed the traditional way by training the model to build the 2D-3D correspondences by 
1) detecting the pre-defined keypoints from image \cite{pavlakos20176,rad2017bb8,tekin2018real,zhou2018starmap} or 
2) predicting the 3D coordinates for object pixels \cite{brachmann2014learning,wang2019normalized,zakharov2019dpod,Li_2019_ICCV}. 
For the former, the sparse 2D-3D correspondences make the estimator sensitive to the occlusion. 
For the latter, RANSAC \cite{nister2005preemptive} is necessary to solve the pose from the dense correspondences, which remarkably increase the computational complexity of the inference. 
Different from these PnP-based approaches, others trained the deep model in an end-to-end manner to predict the pose directly from an image. 
\cite{kehl2017ssd} trained a viewpoint classifier on the SSD \cite{liu2016ssd} by discretizing the 3D rotation space into some bins. 
The translation was calculated from the 2D bounding box. \cite{sundermeyer2018implicit} leveraged an Augmented Auto-Encoder (AAE) to encode features for each discrete rotation to build a dictionary. 
During test, the rotation is retrieved by matching the feature with the dictionary. 
\cite{xiang2017posecnn} developed a CNN-based network to regress the quaternions from image directly. 
These direct approaches can enjoy fast inference speed, but they typically suffer from inferior performance when compared to PnP-based approaches.

Current RGB-based pose estimation approaches rely heavily on real-world annotated training data. 
For instance, with the aid of real data, the pose estimation precision of DPOD \cite{zakharov2019dpod} can be improved from 50\% to 82.98\% on metric ADD. 
Without real training samples, the performance of current approaches is far from satisfactory.
Unfortunately, the pose annotations of real data are quite expensive and time-consuming. 
We propose Pose-Free Reinforced Learning (PFRL) framework, which can significantly improve the pose estimation accuracy only based on the relatively low-cost 2D real-data annotations (i.e. segmentation masks).

\textbf{6D Object Pose Refinement}.
To further promote the performance of 6D pose refinement, numerous approaches leveraged depth by ICP~\cite{besl1992method}. 
As a result, additional depth sensor is required and may be restricted by its the frame rate, resolution, depth range and light illumination, etc.
For RGB-based pose refinement, a promising way is to train a pose refiner. 
For instance, DeepIM~\cite{li2018deepim} was proposed to compare the rendered objects (accompanied with mask) with the observed ones, which can iteratively predict the relative transformation between the current pose estimation and the ground-truth. 
\cite{manhardt2018deep} and \cite{zakharov2019dpod} developed the similar idea by introducing the parallel branches to extract features. 
These approaches can achieve remarkable pose estimation performance by utilizing the ground-truth pose annotations to provide fully-supervised training signals. 
More recently, \cite{periyasamy2019refining} attempted to train a 6D object pose refiner without real 6D pose labels via an approximated differentiable renderer, however the utilization of the differentiable renderer was non-trivial, since they required an additional representation learning step where real 6D pose annotations were still needed.

\textbf{Reinforcement Learning for Vision Tasks}.
Recently, deep reinforcement learning~\cite{sutton2011reinforcement} has been successfully applied to a wide range of computer vision tasks, such as object detection~\cite{mathe2016reinforcement}, semantic segmentation~\cite{zhou2019context}, visual object tracking~\cite{supancic2017tracking,yun2017action,ren2018deep}, vision-based robotic grasping~\cite{quillen2018deep}, etc.
Several works also try to solve 6D object pose estimation related tasks via reinforcement learning. 
For instance, 
\cite{krull2017poseagent} proposes a policy gradient based method to save the inference budget for an existing 6D object pose estimation system. \cite{sock2019active} selects the strategy of camera movements by policy gradient method, directing the camera's attention to the most uncertain object.

Different from them, we propose a novel approach to learn the 6D pose refinement in a Pose-Free fashion without the need of real-world 6D pose annotations via reinforcement learning. 
By formulating the problem under the reinforcement learning framework, the information from the non-differentiable rasterization-based renderer can be better exploited during optimization. 
It enjoys both Pose-Free 2D annotations and pose refinement prediction capability.
\section{Methodology}
The overview of our framework is depicted in Fig. \ref{rlp}.
In following sections, we first present the formulation of Reinforcement Learning (RL) for solving the 6D pose estimation problem in Sec. \ref{rlsetting}. 
We then introduce a 2D mask-based reward function in Sec. \ref{reward}. 
Further, a composite reinforced optimization is proposed for the task-specific 6D pose estimation problem in Sec. \ref{rlalgo}.
Finally, we discuss our policy network architecture in Sec. \ref{netarc}. 

\subsection{Problem Formulation}\label{rlsetting}
To achieve accurate 6D pose estimation, our core idea is to align the 2D projection of 3D object with its appearance observed in 2D image.
We propose the PFRL framework to tackle the problem by reinforcement learning. 
We hypothesize that the decision made by the agent at each time step will be only based on the current state and not be affected by previous states, which can be formulated as an MDP.
The goal is to maximize the expected sum of future discounted rewards $V^\pi(\mathbf{s})=\mathbb{E}\big[\sum_{k\ge 0}\gamma^k r_k\big]$ \cite{puterman2014markov}, where $\gamma\in[0,1)$ is the discount factor, $r_k=r(\mathbf{s}_k,\mathbf{a}_k)$ is immediate reward at time $k$, $\mathbf{s}_k$ is the state at time $k$ and $\mathbf{a}_k\sim \pi(\cdot|\mathbf{s}_k)$ is the action generated by following some policy $\pi$ conditioned on $\mathbf{s}_k$. The state and action space of our framework are defined as follows:

\textbf{State}: 
The state which encodes the knowledge of the environment should be instrumental for the agent to decide how to rotate and translate the 3D model.
At current time step $k$, the state $s_{k}$ is formed by a rendered RGB image and a projected mask of the 3D object model specified by the current pose, and the observed RGB image as well as the ground-truth 2D bounding box of the object. 
These four images are concatenated together as the input of the policy network. During test, since the ground-truth bounding box is not available, we then exploit it by projecting the object with the initial pose instead. 

\textbf{Action}:
Action is the relative SE(3) transformation the agent performs at each time step, which will affect the state of the environment. 
We exploit the same representation of $\mathbf{\Delta p}$ as in \cite{li2018deepim} for its characteristics of disentangled rotation and translation, i.e., the actions of rotation and translation are not mutually affected. 
The rotation and translation actions are chosen stochastically from the two distributions parameterized by policy network. 
Different from conventional reinforcement learning approaches, we measure a single policy value for the two distributions by concatenating the features of rotation and translation together (Sec. \ref{netarc}).
We evaluate continuous and discrete distributions for the policy model respectively:
\begin{equation}\label{Gau}
\mathbf{a}_R\sim \mathcal{N}(\boldsymbol{\mu}_{R}(\theta),\boldsymbol{\sigma}_{R}^2),\mathbf{a}_t\sim \mathcal{N}(\boldsymbol{\mu}_{t}(\theta),\boldsymbol{\sigma}_{t}^2),
\end{equation}
and
\begin{equation}\label{cat}
\mathbf{a}_R\sim \mathcal{C}(\boldsymbol{\zeta}_{R}(\theta)),\mathbf{a}_t\sim \mathcal{C}(\boldsymbol{\zeta}_{t}(\theta)).
\end{equation}

Eq. \ref{Gau} is the continuous distributions for rotation and translation actions. $\mathcal{N}(\cdot)$ represents multivariate Gaussian distribution. 
$\boldsymbol{\mu}_{R}(\theta)$ and $\boldsymbol{\mu}_{t}(\theta)$ represent the mean values of rotation (i.e. quaternion) and translation from the policy network respectively. 
Accordingly, $\boldsymbol{\sigma}_{R}$ and $\boldsymbol{\sigma}_{t}$ are the variance matrices. 
We assume the dimensions are independent of each other for simplicity and without loss of representation ability, hence both of $\boldsymbol{\sigma}_{R}$ and $\boldsymbol{\sigma}_{t}$ are diagonal matrices. 

Eq. \ref{cat} indicates the discrete distribution for rotation and translation actions. $\mathcal{C}(\cdot)$ represents Categorical distribution, where $\boldsymbol{\zeta}_{R}(\theta)$ and $\boldsymbol{\zeta}_{t}(\theta)$ are the probabilities produced by the policy network.

We handcraft 13 operations for rotation and translation actions, where each of them includes rotating (or translating) 3D model along $\pm x$,$\pm y$,$\pm z$ with a fixed small and large degree (or step) as well as no rotation/translation action. See Sec. 1 of the supplementary material for more details.

Each operation for rotation and translation is encoded with a quaternion and a direction vector. 
The final form sampled from the continuous and discrete distribution is the same.

\subsection{2D Mask-based Reward}
\label{reward}
After rotating and translating the 3D model, the agent needs a reward signal $r_A$ for policy update. Under the Pose-Free situation,  we propose a 2D mask-based reward, which includes three components: IoU (Intersection over Union) Difference Reward $r_I$, Goal Reached Reward $r_G$, and Centralization Reward $r_C$, which are purely computed from the rendered and ground-truth masks in 2D space.

We denote the IoU Difference Reward $r_I$ as:
\begin{equation}\label{iour}
r_I=f_\phi(\text{IoU}_{k+1})-f_\phi(\text{IoU}_k),
\end{equation}
where
\begin{equation}
f_\phi(x)=\left \{  
\begin{array}{lll}
x,& x < X_{\text{thr}} \\  \alpha x^2-\beta x,& x\geq X_{\text{thr}}
\end{array} 
\right..
\end{equation}
Let $M_R$ and $M_G$ represent the rendered mask at current state and the ground-truth mask respectively.
In Eq. \ref{iour} $\text{IoU}_k = \frac{S(M_{R_k}\cap M_G)}{S(M_{R_k}\cup M_G)}$ measures their IoU. 
$f_\phi$ is a mapping function, whose shape is controlled by $\alpha$, $\beta$, and a threshold $X_{\text{thr}}$. 
The motivation is that if the mask of an object without high symmetry has a perfect overlap with the ground-truth mask, it is very likely for the object to be in the actual localization and orientation. 
Therefore the pose estimation problem can be converted into a 2D mask matching problem. 
Our objective is to maximize the IoU at the last image frame. 
Therefore, we design $r_I$ in the form of IoU difference in adjacent frames corresponding to the RL objective: 
maximizing the cumulative reward $\sum^{k_\text{end}}_0 r_k$. 
The mapping function $f_\phi$ is designed to give a larger reward when IoU is closer to 1. 
Sharper changes of reward at frames with large IoUs (e.g., 0.8 to 1.0), which is the most common situation during training, help the agent learn the objective much easier.

In practice, when the IoU reaches some threshold $\text{IoU}_{\text{thr}}$, we expect the environment to give the reward feedback immediately and stop refining. 
Thus we define an additional Goal Reached Reward $r_G$ as: 
\begin{equation}
r_G=\left \{  
\begin{array}{lll}
1, &\text{IoU}_k\geq \text{IoU}_{\text{thr}} \\
0,&\text{IoU}_k<\text{IoU}_{\text{thr}}
\end{array} 
\right..
\end{equation}
Such reward gives the agent an evident arrival signal, which saves the budget in real-world applications.

In case of the object locating far away from the true position, which might happen if the network is not well initialized, we further add a Centralization Reward $r_C$ to give an explicit constraint on the estimate of translation. $r_C$ is denoted as:
\begin{equation}\label{ctr}
r_C=\min(||c_r-c_g||_2^{-\frac{1}{2}}, 1),
\end{equation}
where $c_r$ and $c_g$ are the center of the rendered mask and the ground-truth mask, respectively. 

Above all, the final reward $r_A$ can be summarized as: 
\begin{equation}\label{ra}
r_A=r_I+\sigma_C r_C+\sigma_G r_G,
\end{equation}
where $\sigma_C$ and $\sigma_G$ are weights of $r_C$ and $r_G$.

Note that if an object exhibits some symmetry, naively training a pose estimator is problematic and unstable since the similar appearances in RGB can show different 6D poses and lead to distinct loss values \cite{manhardt2019explaining}.
Our design of mask-based reward can implicitly avoid this problem and work well on symmetric objects, similar to the idea in \cite{sundermeyer2018implicit}.

\subsection{Composite Reinforced Optimization}
\label{rlalgo}
The high dimensional state space brings high variance and instability for training the policy network. In usual, millions of samples are required to fully exploit the power of reinforcement learning model. However, in the 6D pose estimation, it is difficult to render such tremendous amount of the images.
In this section, we propose a task-specific composite reinforced optimization method for 6D pose estimation. We combine the on-policy and off-policy strategies together to fully utilize the rendered images.

We use $\theta$ and $\phi$ to denote the current policy network and value function parameters. Given the tuple $(\mathbf{s}_k, \mathbf{a}_k, r_k)$, we use $\theta_{\text{old}}$ to denote the parameters of the policy network where $\mathbf{a}_k$ samples before $\theta$. 
The corresponding policies are denoted as $\pi_\theta(\mathbf{a}_k|\mathbf{s}_k)$ and $\pi_{ \theta_{\text{old}} } (\mathbf{a}_k|\mathbf{s}_k)$. The on-policy optimization refers to learning the value $V_\phi(\mathbf{s}_k)$ of the policy $\pi_\theta(\mathbf{a}_k|\mathbf{s}_k)$ being carried out by the agent with corresponding $\mathbf{a}_k$. In this paper, we employ the Proximal Policy Optimization (PPO) algorithm \cite{schulman2017proximal} for the on-policy optimization. The loss function of PPO is defined as:
\begin{equation}\label{ppoloss}
    L_{\text{on}} = L_{\text{clip}}+\lambda_v  L_{\text{value}} + \lambda_e  L_{\text{entropy}}.
\end{equation}
In Eq. \ref{ppoloss}, $L_{ \text{clip} }$ is the clipped surrogate objective measuring minimum of clipped and original importance-weighted advantage value function. $\lambda_v$ and $\lambda_e$ are trade-off parameters. $L_{\text{value}}$ measures an approximated squared-error loss between value function and cumulative reward. $L_{\text{entropy}}$ is the entropy bonus of policy. 
For further details of the loss function, we refer readers to \cite{schulman2017proximal}.

 After updating policy network, the tuple $(\mathbf{s}_k, \mathbf{a}_k, r_k)$ are no longer associated with the $\pi_\theta$, i.e., $\mathbf{a}_k$ was sampled by a previous policy network $\pi_{\theta_\text{old}}$. 
Therefore the tuple $(\mathbf{s}_k, \mathbf{a}_k, r_k)$ can not be used for on-policy optimization. 
In order to further fully utilize ``outdated'' data samples and take advantage of data efficiency, we introduce an off-policy value update strategy similar to \cite{espeholt2018impala} to assist the on-policy optimization to accelerate the training process. 
We set up a priority queue replay buffer to store the ``outdated'' data samples. 
We then update the value function $V_\phi(\mathbf{s}_k)$ with off-policy loss using samples from the replay buffer:
\begin{equation}\label{loff}
     L_{\text{off}}=(V_\phi(\mathbf{s}_k)-V_{\text{trace}})^2.
\end{equation}
Definitions of $V_{\text{trace}}$ in Eq. \ref{loff} can be found in \cite{espeholt2018impala}. The overall procedure of the composite reinforced optimization is summarized  Algorithm \ref{algoo}. 
\begin{algorithm}[htb] 
    \SetAlgoLined
    Initialize policy network $\pi_\theta$ and value function $V_\phi$\;
    Initialize replay buffer $R$. \\
    \For {$\textup{episode} = 1, M$} {
        Generate trajectories $T=(\mathbf{s}_0, \mathbf{a}_0, r_0, \mathbf{s}_1,...)$\;
        Update $\pi_\theta$ and $V_\phi$ with $L_{\text{on}}$ using $T$\;
        Store $T$ into $R$\;
        Sample $B=(\mathbf{s}, \mathbf{a}, r)$ pairs from $R$\;
        Compute V-trace target $V_\text{trace}$ using $B$\;
        Update $V_\phi$ with $L_\text{off}$. \\
    }
\caption{\small Composite Reinforced Optimization}
\label{algoo}
\end{algorithm}

\subsection{Disentangled Policy Architecture}
\label{netarc}
\begin{figure}[!htbp]
\begin{center}
\includegraphics[height=5.9cm]{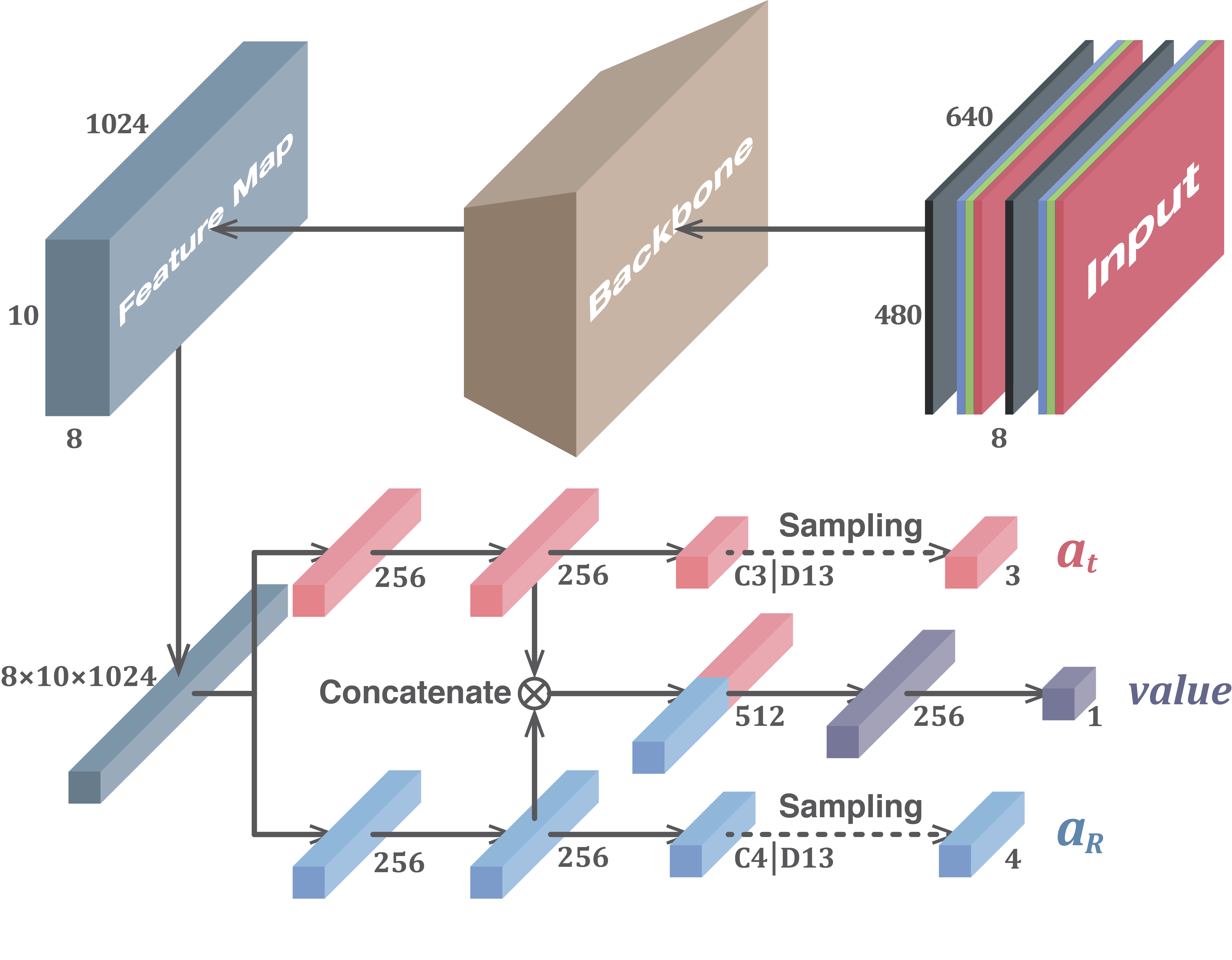}
\end{center}
   \caption{Disentangled Policy Architecture. 
   The observed image, ground-truth bounding box, rendered image, rendered mask are concatenated and zoomed together as input. Branches for rotation and translation are separated and actions are sampled from continuous or discrete distributions. `C3' and `C4' denote the continuous distribution for translation and rotation while the double `D13' denote the discrete distribution. The value function is computed together.}
\label{network}
\end{figure}
Fig. \ref{network} depicts the disentangled policy network structure. 
The backbone consists of the first 11 convolutional layers of FlowNet-S \cite{dosovitskiy2015flownet}. 
Similar to \cite{li2018deepim}, the input images and masks are zoomed-in before feeding to the backbone.
As discussed in Sec. \ref{rlsetting}, two separate branches are used to predict the distribution of the disentangled rotation and translation. 
Both branches contain 2 fully-connected (FC) layers with 256 hidden units, and are followed by one additional FC layer. 
Compared with \cite{li2018deepim}, our network is lightweight both in training and inference, since we do not use the heavily computational flow and mask branches. 
The last FC layers of both branches are concatenated together followed by 2 FC layers to produce the value function. 
In this manner, the value function shares the same parameters as actions, and can be used for optimization of both actions. 
\section{Experiments}
\begin{table*}[htbp]

\begin{center}
\resizebox{\textwidth}{29mm}{
\begin{tabular}{r|cc|ccc||cc|cc|cc}
\toprule[2pt]
\textbf{Train data} & \multicolumn{2}{c|}{\textbf{Pose-Free Init}} & \multicolumn{3}{c||}{\textbf{+Pose-Free Refine}} & \multicolumn{2}{c|}{\textbf{Gt Pose Init}}&\multicolumn{4}{c}{\textbf{+Pose-Free Refine}}\\
\midrule[1pt]
\multirow{2}{*}{\textbf{Object}}&\textbf{AAE}\cite{sundermeyer2018implicit}&\textbf{DPOD-SYN}\cite{zakharov2019dpod}&\textbf{SSD6D}\cite{kehl2017ssd}&\textbf{DPOD-SYN+Refine}&\textbf{AAE+ours}&\textbf{YOLO6D}\cite{tekin2018real}&\textbf{PoseCNN}\cite{xiang2017posecnn}&\multicolumn{2}{c|}{\textbf{PoseCNN+DeepIM-SYN}\cite{li2018deepim}}&\multicolumn{2}{c}{\textbf{PoseCNN+ours}}\\
&\textbf{ADD}&\textbf{ADD}&\textbf{ADD}&\textbf{ADD}&\textbf{ADD}&\textbf{ADD}&\textbf{ADD}&\textbf{Proj.2D}&\textbf{ADD}&\textbf{Proj.2D}&\textbf{ADD}\\
\midrule[1pt]
Ape        & 3.96&37.22&-&55.23&\textbf{65.4}&21.6&27.8&81.7&23.9&\textbf{95.6}&\textbf{60.5}\\
Benchvise  &20.92&66.76&-&72.69&\textbf{84.5}&81.8&68.9&92.2&\textbf{93.1}&\textbf{94.7}&88.9\\
Camera     &30.47&24.22&-&34.76&\textbf{41.5}&36.6&47.5&\textbf{97.0}&\textbf{84.7}&95.0&64.6\\
Can        &35.87&52.57&-&\textbf{83.59}&80.9&68.8&71.4&89.6&\textbf{91.5}&\textbf{93.1}&91.3\\
Cat        &17.90&32.36&-&65.10&\textbf{80.4}&41.8&56.7&96.1&79.5&\textbf{99.3}&\textbf{82.9}\\
Driller    &23.99&66.60&-&73.32&\textbf{77.6}&63.5&65.4&85.9&82.3&\textbf{94.8}&\textbf{92.0}\\
Duck       & 4.86&26.12&-&50.04&\textbf{52.5}&27.2&42.8&92.6&24.0&\textbf{98.2}&\textbf{55.2}\\
Eggbox     &81.01&73.35&-&89.05&\textbf{96.1}&69.6&98.3&90.8&88.3&\textbf{97.8}&\textbf{99.4}\\
Glue       &45.49&74.96&-&\textbf{84.37}&76.7&80.0&95.6&81.2&\textbf{96.9}&\textbf{97.1}&93.3\\
Holepuncher&17.60&24.50&-&35.35&\textbf{44.9}&42.6&50.9&78.0&20.6&\textbf{96.7}&\textbf{66.7}\\
Iron       &32.03&85.02&-&\textbf{98.78}&67.3&75.0&65.6&59.3&\textbf{85.1}&\textbf{81.6}&75.8\\
Lamp       &60.47&57.26&-&74.27&\textbf{91.1}&71.1&70.3&75.6&85.5&\textbf{96.0}&\textbf{96.6}\\
Phone      &33.79&29.08&-&46.98&\textbf{52.7}&47.7&54.6&88.3&66.1&\textbf{91.0}&\textbf{69.1}\\
\midrule[1pt]
Mean      & 31.41 & 50.0 & 34.1 & 66.43 & \textbf{70.1} & 56.0 & 62.7 & 85.3 & 70.9 & \textbf{94.7} & \textbf{79.7}\\
\bottomrule[2pt]
\end{tabular}}
\end{center}
\caption{Comparison with state-of-the-art Pose-Free methods on LINEMOD with metrics ADD and Proj.~2D. 
Left part is results of initial pose trained on synthetic data + Pose-Free refiner, and right part is results of initial pose trained on ground-truth labels + Pose-Free refiner. 
PFRL (ours) outperforms the state-of-the-art method DPOD trained with synthetic data, despite that we use a much worse initialization method provided by AAE. 
When using the same initial pose from PoseCNN, PFRL outperforms DeepIM trained on synthetic data.}
\label{LM6dR}
\end{table*}
\subsection{Data Preparation}
We conduct our experiments on LINEMOD \cite{hinterstoisser2012model} and T-LESS \cite{hodan2017t} dataset. 
We split the LINEMOD dataset following \cite{brachmann2016uncertainty}, including about 15\% for training and 85\% for test, i.e., around 200 images per object for the training set. 
Note that no 6D pose annotations are directly used during training and testing. They are only used to generate ground-truth masks. Under the case that 6D pose is hard or impossible to obtain, we can get the ground-truth mask by semantic segmentation or manual annotation.

On LINEMOD dataset, we use results from AAE~\cite{sundermeyer2018implicit} and PoseCNN~\cite{xiang2017posecnn} as our initial pose for test. 
Random Gaussian noise with zero mean is added to each dimension of the rotation and translation as initial pose for training.  
The variance is $(15^\circ)^2$ for each rotation axis and $[2^2,2^2,5^2](cm)^2$ for translation $[x,y,z]$. 
A noise is resampled if one of the rotation axis exceeds 45 degrees.

On T-LESS dataset we initialize our method with results from AAE~\cite{sundermeyer2018implicit}, the current state-of-the-art RGB-only method on T-LESS. 
No real images are available for training, so we render objects with random poses and light conditions on random background images from PASCAL VOC~\cite{everingham2010pascal}. We perturb the ground-truth poses with the same noise used for LINEMOD as initial poses. 
Since the camera intrinsic matrix of each real image differs only in the principal points slightly, 
we translate and crop the images such that they have the same intrinsic matrix for ease of use.
\subsection{Evaluation Metrics}
In our experiments, we use three common metrics for evaluation: Proj.~2D, ADD and VSD. 
For Proj.~2D metric, a pose is regarded as correct if the average Proj.~2D distance (Eq.(\ref{2dproj})) of the model points is less than 5 pixels:
\begin{equation}\label{2dproj}
\textbf{Proj.~2D}=\frac{1}{m} \sum_{x \in \mathcal{M}}\|\mathbf{K}(\mathbf{R} \mathbf{x}+\mathbf{t})-\mathbf{K}(\hat{\mathbf{R}} \mathbf{x}+\hat{\mathbf{t}})\|.
\end{equation}
For ADD metric, a pose is regarded as correct if the average distance of model points is less than 10\% of the model diameter~\cite{hinterstoisser2012model} (Eq.(\ref{add1})). 
For symmetric objects, the distance is calculated by the closest model point (Eq.(\ref{adi})). 
\begin{equation}
\label{add1}
\textbf{ADD}=\frac{1}{m} \sum_{x \in \mathcal{M}}\|(\mathbf{R} \mathbf{x}+\mathbf{t})-(\hat{\mathbf{R}} \mathbf{x}+\hat{\mathbf{t}})\|,
\end{equation}
\begin{equation}
\label{adi}
\textbf{ADD-S}=\frac{1}{m} \sum_{\mathbf{x}_1 \in \mathcal{M}} \min _{\mathbf{x}_2 \in \mathcal{M}}\|(\mathbf{R} \mathbf{x}_1+\mathbf{t})-(\hat{\mathbf{R}} \mathbf{x}_2+\hat{\mathbf{t}})\|.
\end{equation}
Visible Surface Discrepancy (VSD)~\cite{hodavn2016evaluation} is an ambiguity-invariant metric that depends on the visible surface, which can account for the symmetry better. The parameters used for VSD metric are the same as \cite{sundermeyer2018implicit}.
\begin{figure*}[htbp]
\begin{center}
\includegraphics[height=3.6cm]{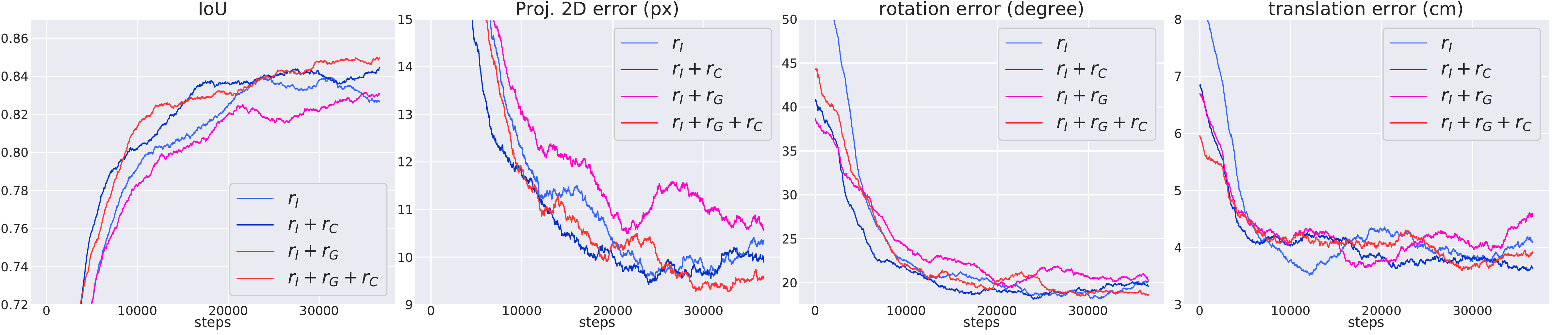}
\end{center}
\caption{Curve of different reward styles.}
\label{rewardcurve}
\end{figure*}
\subsection{Results on LINEMOD}

Table \ref{LM6dR} shows the results of our method and some recent well-performed RGB-based methods. 
We achieve state-of-the-art on Pose-Free methods. 
Most Pose-Free methods like DPOD~\cite{zakharov2019dpod} and AAE~\cite{sundermeyer2018implicit} render objects on random backgrounds such as images from PASCAL VOC dataset~\cite{everingham2010pascal} and train on synthetic data. 
When the initial pose estimation and the refiner are both Pose-Free, our method achieves average ADD 70.1\% with initial pose estimated by AAE (ADD 31.41\%), surpassing the previous state-of-the art method DPOD's ADD 66.43\% with initial pose ADD 50.0\%. 
And on 10 of 13 objects we get a higher score than DPOD despite the relatively lower initial pose score.
We bring on average 38.7 percentages ADD improvement on AAE.

To evaluate our method with better initial poses, we also use the results of PoseCNN~\cite{xiang2017posecnn} trained with real 6D pose labels as initial poses. 
We bring 17.0 percentages improvement on PoseCNN. 
Our method also surpasses DeepIM-SYN~\cite{li2018deepim}, the DeepIM trained with pure synthetic data and initialized with PoseCNN results during test, on both metrics overall. 
We can also see that on Proj.~2D metric, our method performs better on 12 of 13 objects, which demonstrates the advantage of our design of the 2D mask based reward. We also provide results with DPOD initialization in Sec. 3 of the supplementary material.

\subsection{Results on T-LESS}
To evaluate our method on synthetic-only cases, we train on objects 19-23 of T-LESS dataset. 
We use AAE~\cite{sundermeyer2018implicit} with ground-truth bounding box as our initial pose for testing. 
Objects 19-23 are the only objects trained with texture-less models for AAE.
Since T-LESS dataset has no real images for training, we use synthetic data for training similar to AAE. 
The results are shown in Table \ref{tless}. 

\begin{table}[htbp]
\begin{center}
\begin{tabular}{r|cc|cc}
\toprule[2pt]
\textbf{Method}&\multicolumn{2}{c|}{\textbf{AAE}\cite{sundermeyer2018implicit}}&\multicolumn{2}{c}{\textbf{AAE+Ours}}\\
\textbf{Metric}&\textbf{Proj.~2D}&\textbf{VSD}&\textbf{Proj.~2D}&\textbf{VSD}\\
\midrule[1pt]
19&30.65&49.95&32.79&57.39\\
20&23.51&41.87&25.40&45.29\\
21&56.55&59.06&60.58&62.50\\
22&42.99&46.08&44.78&48.02\\
23&21.88&40.38&29.32&44.44\\
\midrule[1pt]
Mean&35.12&47.47&\textbf{38.57}&\textbf{51.53}\\
\bottomrule[2pt]
\end{tabular}
\end{center}
\caption{Results on the T-LESS dataset. Accuracy in the left side is calculated from the models provided by the author of AAE \cite{sundermeyer2018implicit} and is slightly better than the original reported results.}
\label{tless}
\end{table}
On texture-less objects without real training images, the improvement of metric recall is not as obvious as in LINEMOD due to the large domain gap.
However we still get 3.5 percentages improvement on Proj.~2D and 4.1 percentages on the more widely used metric VSD. 
We believe further improvement can be achieved with the availability of real images with mask annotations, which is much cheaper to obtain than 6D pose annotations.

\subsection{Ablation study}
We do ablation study on the reward style, action style, testing speed and optimization strategy. 
For each ablation we use several objects from LINEMOD \cite{hinterstoisser2012model} for training and testing. 
We set batch size to 256, $X_\text{thr}=0.5$, $\text{IoU}_\text{thr}=0.98$, $\alpha=3.8$, $\beta=1.8$, $\sigma_C = 1$, $\sigma_G=2$, $\lambda_v=0.5$, $\lambda_e=0.001$.
All hyper-parameters are fixed unless otherwise stated. 

\textbf{Reward style}. 
Our basic reward is the IoU Difference Reward $r_I$. 
We train on object Iron with the same parameters and training steps using $r_I$ and the additional Goal Reached Reward $r_G$ and the Centralization Reward $r_C$. 
Fig. \ref{rewardcurve} shows the comparison of four reward styles in four evaluation metrics. 
Although we can not see obvious difference from the average rotation and translation error since rewards are designed from the mask information, the IoU of $r_I+r_G+r_C$ increases fastest and maintains the highest level, while it has the lowest Proj.~2D error. 
Comparing $r_I+r_C$ with $r_I$, $r_C$ brings obvious acceleration on the IoU improvement. 
Although adding $r_G$ to $r_I$ independently seems bad, it performs well when combined with $r_C$. 
Therefore in the rest part of our experiment, we all use reward $r_I+r_G+r_C$.

\begin{figure*}[htbp]
\begin{center}
\includegraphics[width=18cm]{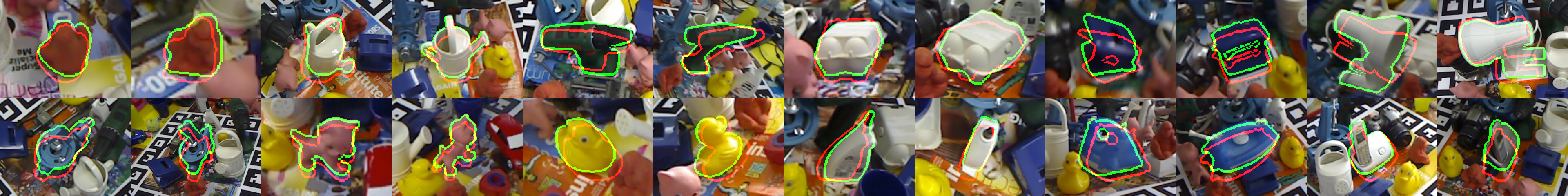}
\end{center}
   \caption{Qualitative results of AAE initial poses~\cite{sundermeyer2018implicit} and our refined poses on the LINEMOD dataset. 
   The red and green lines represent the contours of the initial poses and our refined poses respectively.}
\label{mask_r}
\end{figure*}
\begin{table}[htbp]
\begin{center}
\begin{tabular}{r|cc|cc}
\toprule[2pt]
\textbf{Action Style}&\multicolumn{2}{c|}{\textbf{Continuous}}&\multicolumn{2}{c}{\textbf{Discrete}}\\
\textbf{Metric}&\textbf{Proj.~2D}&\textbf{ADD}&\textbf{Proj.~2D}&\textbf{ADD}\\
\midrule[1pt]
Initial   &14.2&35.6&\textbf{14.2}&\textbf{35.6}\\
Epoch:  800&37.6&43.8&\textbf{58.9}&\textbf{52.8}\\
Epoch:1600&45.7&46.7&\textbf{63.3}&\textbf{55.4}\\
Epoch:2400&57.5&56.2&\textbf{75.3}&\textbf{70.7}\\
Epoch:3200&60.4&60.9&\textbf{80.6}&\textbf{79.2}\\
\bottomrule[2pt]
\end{tabular}
\end{center}
\caption{Ablation on action style.}
\label{acst}
\end{table}
\textbf{Action style.}
We train 2 models on the same conditions except that the action styles are continuous and discrete separately and test on the Benchvise object. 
Table \ref{acst} shows the results for different training epochs on the test set. 
We can tell that discrete action space outperforms continuous one for the whole training procedure, and performs around 20\% better in convergence. 
Discretization makes great contribution to our method.
\begin{figure}[htbp]
\begin{center}
\includegraphics[height=4cm]{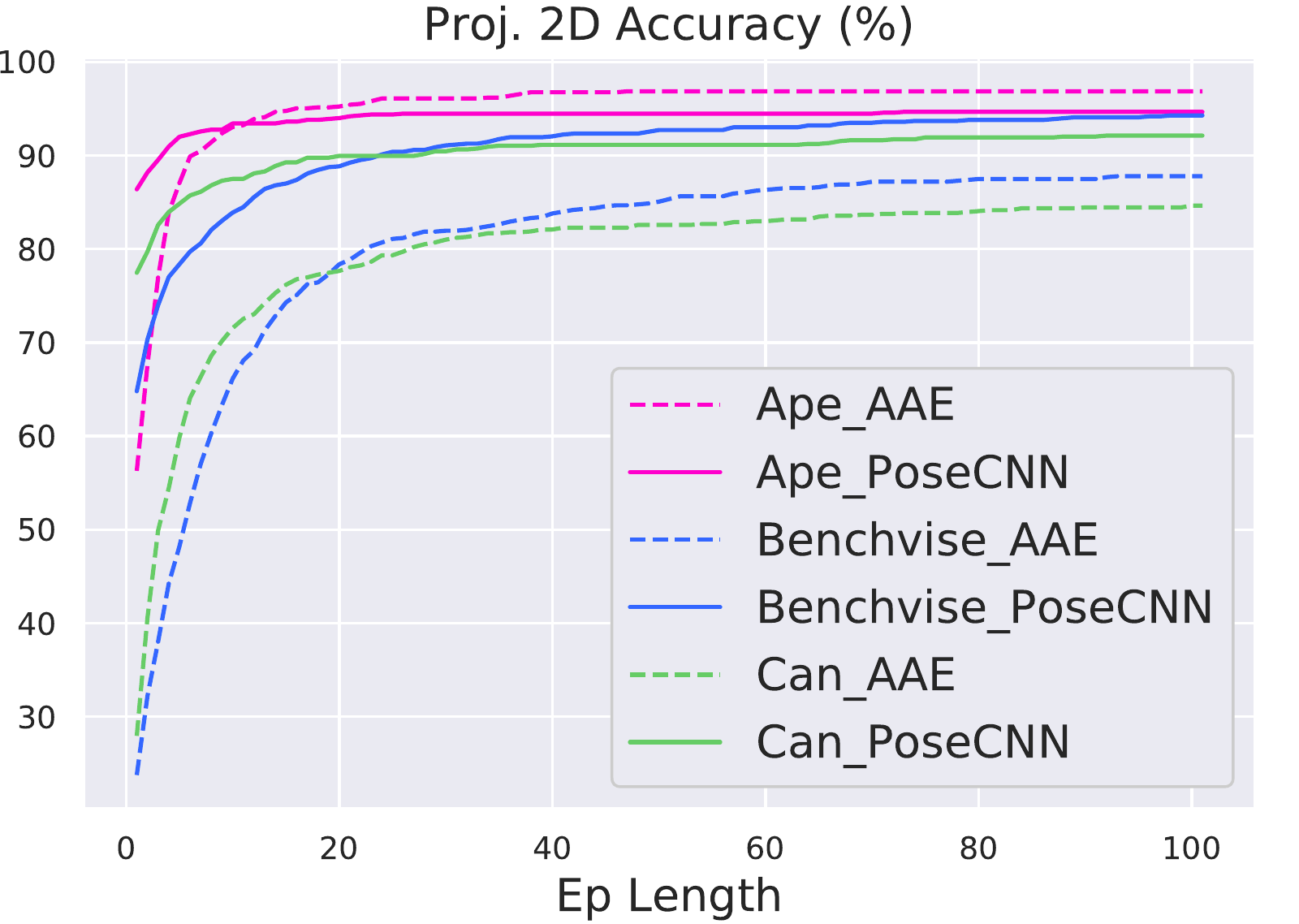}
\end{center}
   \caption{The Proj.~2D accuracy with respect to test episode length curves of 3 objects: Ape, Benchvise, Can.}
\label{epl}
\end{figure}

\textbf{Effect of refining steps and testing speed.}
Our experiments were conducted on a desktop with an Intel Core i7 3.60GHz CPU and a NVIDIA GeForce RTX 2080 Ti GPU. 
We set episode length to 50 for one pose estimation process during training. 
We find that the object usually gets very close to the target pose in a few steps and keeps motionless in the next steps. 
Therefore we plot the curve of Proj.~2D accuracy with respect to refining steps in Fig.~\ref{epl}. 
Curves include 3 objects (Ape, Benchvise, Can) with 2 kinds of initial poses (AAE, PoseCNN). 
We can see that all objects reach their best accuracy at step 15-20. 
We summarize the accuracy and test speed(ms per frame) with respect to the episode length using AAE~\cite{sundermeyer2018implicit} as initial pose in Table~\ref{eplt}. 
\begin{table}[htbp]
\begin{center}

\begin{tabular}{r|c|C{0.6cm}C{0.6cm}|C{0.6cm}C{0.6cm}|C{0.6cm}C{0.6cm}}
\toprule[2pt]
\multicolumn{2}{c|}{\multirow{2}{*}{\textbf{Ep Length}}}&\multicolumn{2}{c|}{\textbf{Ape}}&\multicolumn{2}{c|}{\textbf{Benchvise}}&\multicolumn{2}{c}{\textbf{Can}}\\
\multicolumn{2}{c|}{}&\textbf{A}&\textbf{P}&\textbf{A}&\textbf{P}&\textbf{A}&\textbf{P}\\
\midrule[1pt]
 \multirow{5}{*}{\makecell[r]{Acc\\(\%)}} &0 &39.0&83.1&14.2&50.0&16.8&69.6\\
 						  &5 &87.0&92.0&48.2&78.4&59.8&84.8\\
 						  &10&93.0&93.4&66.1&83.9&71.6&87.5\\
 						  &20&95.2&94.0&78.4&88.9&77.7&90.0\\
 						  &50&96.9&94.2&85.1&92.7&82.6&91.1\\
 \midrule[1pt]
 \multirow{4}{*}{\makecell[r]{Time\\(ms)}}
 						  &5 &62 &64 &67 &65 &68 &68\\
 						  &10&110&114&126&121&128&127\\
 						  &20&218&220&243&234&245&243\\
 						  &50&529&534&588&578&588&584\\ 
\bottomrule[2pt]
\end{tabular}
\end{center}
\caption{Accuracy and testing time with respect to episode length. `A' and `P' denote initial poses from AAE~\cite{sundermeyer2018implicit} and PoseCNN~\cite{xiang2017posecnn}, respectively.}
\label{eplt}
\end{table}
Our method is flexible in terms of accuracy and refining steps. 
We find that a relatively good pose can be estimated in 20 steps, that is, in about 240ms per frame.
\begin{figure}[htbp]
\begin{center}
\includegraphics[height=3.4cm]{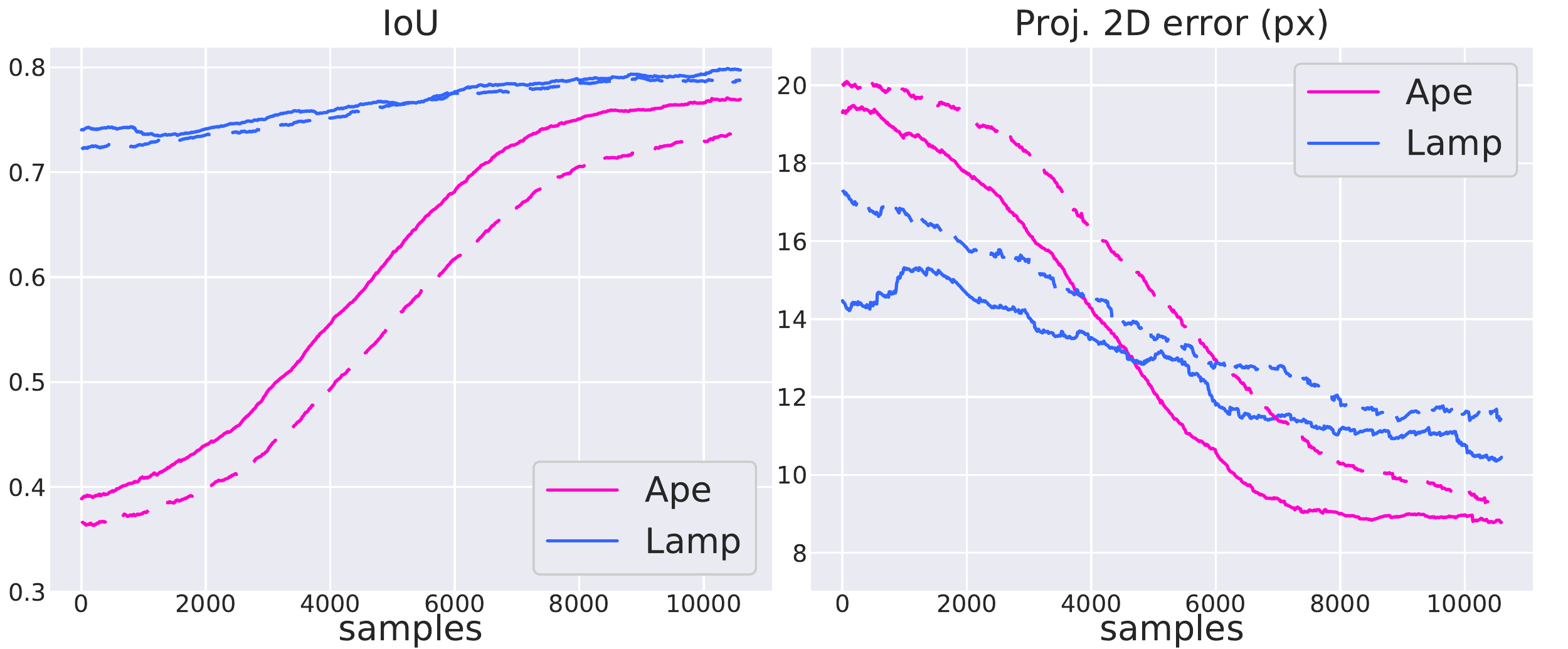}
\end{center}
   \caption{Composite optimization strategy (solid line) versus on-policy optimization strategy (dashed line).}
\label{algoc}
\end{figure}

\textbf{Optimization strategy.} 
We compare the composite optimization strategy with the pure on-policy optimization strategy on objects Ape and Lamp under the same number of samples in Fig.~\ref{algoc}.
The 4 curves in each figure represents 2 objects with 2 optimization strategies separately. 
We use the same learning rate $10^{-4}$ for on-policy and off-policy update. 
The on-policy and off-policy update times are 4:1. 
The left and right shows the change of IoU and Proj.~2D error with the number of updated samples. 
We can tell that for the same object and the same update samples, mixed update rule has higher IoU and lower Proj.~2D error than on-policy update rule, which demonstrates that
 extra updates on value functions help the policy get an accurate value earlier without more samples acquired from the environment.
 
 For more experiment results such as generalization ability and class-agnostic training, we kindly refer readers to supplementary material.
\section{Conclusion}
In this work we formulate the RGB-based 6D pose estimation problem as an MDP and introduce PFRL framework in a Pose-Free fashion without the need of real-world 6D pose annotations. 
We design a task-specified 2D mask-based reward which is purely computed from the object mask information and employ a composite reinforced optimization rule to learn the operation policy efficiently and effectively. 
The experiments demonstrate that our approach is able to achieve the state-of-the-art performance compared with the methods without using real-world ground-truth 6D pose labels on LINEMOD and T-LESS dataset.
\section{Acknowledgement}
This work was supported by the National Key R$\&$D Program of China under Grant 2018AAA0102800 and 2018AAA0102801.
{\small
\bibliographystyle{ieee_fullname}
\bibliography{arxiv}

\begin{thebibliography}{10}\itemsep=-1pt

\bibitem{besl1992method}
Paul~J Besl and Neil~D McKay.
\newblock Method for registration of 3-d shapes.
\newblock In {\em Sensor fusion IV: control paradigms and data structures},
  volume 1611, pages 586--606. International Society for Optics and Photonics,
  1992.

\bibitem{brachmann2014learning}
Eric Brachmann, Alexander Krull, Frank Michel, Stefan Gumhold, Jamie Shotton,
  and Carsten Rother.
\newblock Learning 6d object pose estimation using 3d object coordinates.
\newblock In {\em European conference on computer vision}, pages 536--551.
  Springer, 2014.

\bibitem{brachmann2016uncertainty}
Eric Brachmann, Frank Michel, Alexander Krull, Michael Ying~Yang, Stefan
  Gumhold, et~al.
\newblock Uncertainty-driven 6d pose estimation of objects and scenes from a
  single rgb image.
\newblock In {\em Proceedings of the IEEE Conference on Computer Vision and
  Pattern Recognition}, pages 3364--3372, 2016.

\bibitem{chen2017multi}
Xiaozhi Chen, Huimin Ma, Ji Wan, Bo Li, and Tian Xia.
\newblock Multi-view 3d object detection network for autonomous driving.
\newblock In {\em Proceedings of the IEEE Conference on Computer Vision and
  Pattern Recognition}, pages 1907--1915, 2017.

\bibitem{collet2011moped}
Alvaro Collet, Manuel Martinez, and Siddhartha~S Srinivasa.
\newblock The moped framework: Object recognition and pose estimation for
  manipulation.
\newblock {\em The International Journal of Robotics Research},
  30(10):1284--1306, 2011.

\bibitem{dosovitskiy2015flownet}
Alexey Dosovitskiy, Philipp Fischer, Eddy Ilg, Philip Hausser, Caner Hazirbas,
  Vladimir Golkov, Patrick Van Der~Smagt, Daniel Cremers, and Thomas Brox.
\newblock Flownet: Learning optical flow with convolutional networks.
\newblock In {\em Proceedings of the IEEE international conference on computer
  vision}, pages 2758--2766, 2015.

\bibitem{espeholt2018impala}
Lasse Espeholt, Hubert Soyer, Remi Munos, Karen Simonyan, Volodymyr Mnih, Tom
  Ward, Yotam Doron, Vlad Firoiu, Tim Harley, Iain Dunning, et~al.
\newblock Impala: Scalable distributed deep-rl with importance weighted
  actor-learner architectures.
\newblock In {\em International Conference on Machine Learning}, pages
  1406--1415, 2018.

\bibitem{everingham2010pascal}
Mark Everingham, Luc Van~Gool, Christopher~KI Williams, John Winn, and Andrew
  Zisserman.
\newblock The pascal visual object classes (voc) challenge.
\newblock {\em International journal of computer vision}, 88(2):303--338, 2010.

\bibitem{hinterstoisser2012model}
Stefan Hinterstoisser, Vincent Lepetit, Slobodan Ilic, Stefan Holzer, Gary
  Bradski, Kurt Konolige, and Nassir Navab.
\newblock Model based training, detection and pose estimation of texture-less
  3d objects in heavily cluttered scenes.
\newblock In {\em Asian conference on computer vision}, pages 548--562.
  Springer, 2012.

\bibitem{hodan2017t}
Tom{\'a}{\v{s}} Hodan, Pavel Haluza, {\v{S}}tep{\'a}n Obdr{\v{z}}{\'a}lek, Jiri
  Matas, Manolis Lourakis, and Xenophon Zabulis.
\newblock T-less: An rgb-d dataset for 6d pose estimation of texture-less
  objects.
\newblock In {\em 2017 IEEE Winter Conference on Applications of Computer
  Vision (WACV)}, pages 880--888. IEEE, 2017.

\bibitem{hodavn2016evaluation}
Tom{\'a}{\v{s}} Hoda{\v{n}}, Ji{\v{r}}{\'\i} Matas, and {\v{S}}t{\v{e}}p{\'a}n
  Obdr{\v{z}}{\'a}lek.
\newblock On evaluation of 6d object pose estimation.
\newblock In {\em European Conference on Computer Vision}, pages 606--619.
  Springer, 2016.

\bibitem{kehl2017ssd}
Wadim Kehl, Fabian Manhardt, Federico Tombari, Slobodan Ilic, and Nassir Navab.
\newblock Ssd-6d: Making rgb-based 3d detection and 6d pose estimation great
  again.
\newblock In {\em Proceedings of the IEEE International Conference on Computer
  Vision}, pages 1521--1529, 2017.

\bibitem{krull2017poseagent}
Alexander Krull, Eric Brachmann, Sebastian Nowozin, Frank Michel, Jamie
  Shotton, and Carsten Rother.
\newblock Poseagent: Budget-constrained 6d object pose estimation via
  reinforcement learning.
\newblock In {\em Proceedings of the IEEE Conference on Computer Vision and
  Pattern Recognition}, pages 6702--6710, 2017.

\bibitem{lepetit2009epnp}
Vincent Lepetit, Francesc Moreno-Noguer, and Pascal Fua.
\newblock Epnp: An accurate o (n) solution to the pnp problem.
\newblock {\em International journal of computer vision}, 81(2):155, 2009.

\bibitem{li2018deepim}
Yi Li, Gu Wang, Xiangyang Ji, Yu Xiang, and Dieter Fox.
\newblock Deepim: Deep iterative matching for 6d pose estimation.
\newblock In {\em Proceedings of the European Conference on Computer Vision
  (ECCV)}, pages 683--698, 2018.

\bibitem{Li_2019_ICCV}
Zhigang Li, Gu Wang, and Xiangyang Ji.
\newblock Cdpn: Coordinates-based disentangled pose network for real-time
  rgb-based 6-dof object pose estimation.
\newblock In {\em The IEEE International Conference on Computer Vision (ICCV)},
  October 2019.

\bibitem{liu2016ssd}
Wei Liu, Dragomir Anguelov, Dumitru Erhan, Christian Szegedy, Scott Reed,
  Cheng-Yang Fu, and Alexander~C Berg.
\newblock Ssd: Single shot multibox detector.
\newblock In {\em European conference on computer vision}, pages 21--37.
  Springer, 2016.

\bibitem{manhardt2019explaining}
Fabian Manhardt, Diego~Martin Arroyo, Christian Rupprecht, Benjamin Busam,
  Tolga Birdal, Nassir Navab, and Federico Tombari.
\newblock Explaining the ambiguity of object detection and 6d pose from visual
  data.
\newblock In {\em Proceedings of the IEEE International Conference on Computer
  Vision}, pages 6841--6850, 2019.

\bibitem{manhardt2019roi}
Fabian Manhardt, Wadim Kehl, and Adrien Gaidon.
\newblock Roi-10d: Monocular lifting of 2d detection to 6d pose and metric
  shape.
\newblock In {\em Proceedings of the IEEE Conference on Computer Vision and
  Pattern Recognition}, pages 2069--2078, 2019.

\bibitem{manhardt2018deep}
Fabian Manhardt, Wadim Kehl, Nassir Navab, and Federico Tombari.
\newblock Deep model-based 6d pose refinement in rgb.
\newblock In {\em Proceedings of the European Conference on Computer Vision
  (ECCV)}, pages 800--815, 2018.

\bibitem{marchand2015pose}
Eric Marchand, Hideaki Uchiyama, and Fabien Spindler.
\newblock Pose estimation for augmented reality: a hands-on survey.
\newblock {\em IEEE transactions on visualization and computer graphics},
  22(12):2633--2651, 2015.

\bibitem{mathe2016reinforcement}
Stefan Mathe, Aleksis Pirinen, and Cristian Sminchisescu.
\newblock Reinforcement learning for visual object detection.
\newblock In {\em Proceedings of the IEEE Conference on Computer Vision and
  Pattern Recognition}, pages 2894--2902, 2016.

\bibitem{nister2005preemptive}
David Nist{\'e}r.
\newblock Preemptive ransac for live structure and motion estimation.
\newblock {\em Machine Vision and Applications}, 16(5):321--329, 2005.

\bibitem{pavlakos20176}
Georgios Pavlakos, Xiaowei Zhou, Aaron Chan, Konstantinos~G Derpanis, and
  Kostas Daniilidis.
\newblock 6-dof object pose from semantic keypoints.
\newblock In {\em 2017 IEEE International Conference on Robotics and Automation
  (ICRA)}, pages 2011--2018. IEEE, 2017.

\bibitem{peng2019pvnet}
Sida Peng, Yuan Liu, Qixing Huang, Xiaowei Zhou, and Hujun Bao.
\newblock Pvnet: Pixel-wise voting network for 6dof pose estimation.
\newblock In {\em Proceedings of the IEEE Conference on Computer Vision and
  Pattern Recognition}, pages 4561--4570, 2019.

\bibitem{periyasamy2019refining}
Arul~Selvam Periyasamy, Max Schwarz, and Sven Behnke.
\newblock Refining 6d object pose predictions using abstract
  render-and-compare.
\newblock In {\em IEEE-RAS International Conference on Humanoid Robots
  (Humanoids)}, 2019.

\bibitem{puterman2014markov}
Martin~L Puterman.
\newblock {\em Markov Decision Processes.: Discrete Stochastic Dynamic
  Programming}.
\newblock John Wiley \& Sons, 2014.

\bibitem{quillen2018deep}
Deirdre Quillen, Eric Jang, Ofir Nachum, Chelsea Finn, Julian Ibarz, and Sergey
  Levine.
\newblock Deep reinforcement learning for vision-based robotic grasping: A
  simulated comparative evaluation of off-policy methods.
\newblock In {\em 2018 IEEE International Conference on Robotics and Automation
  (ICRA)}, pages 6284--6291. IEEE, 2018.

\bibitem{rad2017bb8}
Mahdi Rad and Vincent Lepetit.
\newblock Bb8: A scalable, accurate, robust to partial occlusion method for
  predicting the 3d poses of challenging objects without using depth.
\newblock In {\em Proceedings of the IEEE International Conference on Computer
  Vision}, pages 3828--3836, 2017.

\bibitem{ren2018deep}
Liangliang Ren, Xin Yuan, Jiwen Lu, Ming Yang, and Jie Zhou.
\newblock Deep reinforcement learning with iterative shift for visual tracking.
\newblock In {\em Proceedings of the European Conference on Computer Vision
  (ECCV)}, pages 684--700, 2018.

\bibitem{schulman2017proximal}
John Schulman, Filip Wolski, Prafulla Dhariwal, Alec Radford, and Oleg Klimov.
\newblock Proximal policy optimization algorithms.
\newblock {\em arXiv preprint arXiv:1707.06347}, 2017.

\bibitem{sock2019active}
Juil Sock, Guillermo Garcia-Hernando, and Tae-Kyun Kim.
\newblock Active 6d multi-object pose estimation in cluttered scenarios with
  deep reinforcement learning, 2019.

\bibitem{sundermeyer2018implicit}
Martin Sundermeyer, Zoltan-Csaba Marton, Maximilian Durner, Manuel Brucker, and
  Rudolph Triebel.
\newblock Implicit 3d orientation learning for 6d object detection from rgb
  images.
\newblock In {\em Proceedings of the European Conference on Computer Vision
  (ECCV)}, pages 699--715, 2018.

\bibitem{supancic2017tracking}
James Supancic~III and Deva Ramanan.
\newblock Tracking as online decision-making: Learning a policy from streaming
  videos with reinforcement learning.
\newblock In {\em Proceedings of the IEEE International Conference on Computer
  Vision}, pages 322--331, 2017.

\bibitem{sutton2011reinforcement}
Richard~S Sutton and Andrew~G Barto.
\newblock Reinforcement learning: An introduction.
\newblock 2011.

\bibitem{tekin2018real}
Bugra Tekin, Sudipta~N Sinha, and Pascal Fua.
\newblock Real-time seamless single shot 6d object pose prediction.
\newblock In {\em Proceedings of the IEEE Conference on Computer Vision and
  Pattern Recognition}, pages 292--301, 2018.

\bibitem{wang2019normalized}
He Wang, Srinath Sridhar, Jingwei Huang, Julien Valentin, Shuran Song, and
  Leonidas~J Guibas.
\newblock Normalized object coordinate space for category-level 6d object pose
  and size estimation.
\newblock In {\em Proceedings of the IEEE Conference on Computer Vision and
  Pattern Recognition}, pages 2642--2651, 2019.

\bibitem{xiang2017posecnn}
Yu Xiang, Tanner Schmidt, Venkatraman Narayanan, and Dieter Fox.
\newblock {PoseCNN}: A convolutional neural network for 6{D} object pose
  estimation in cluttered scenes.
\newblock {\em Robotics: Science and Systems (RSS)}, 2018.

\bibitem{yun2017action}
Sangdoo Yun, Jongwon Choi, Youngjoon Yoo, Kimin Yun, and Jin Young~Choi.
\newblock Action-decision networks for visual tracking with deep reinforcement
  learning.
\newblock In {\em Proceedings of the IEEE conference on computer vision and
  pattern recognition}, pages 2711--2720, 2017.

\bibitem{zakharov2019dpod}
Sergey Zakharov, Ivan Shugurov, and Slobodan Ilic.
\newblock Dpod: 6d pose object detector and refiner.
\newblock In {\em Proceedings of the IEEE International Conference on Computer
  Vision}, pages 1941--1950, 2019.

\bibitem{zhou2018starmap}
Xingyi Zhou, Arjun Karpur, Linjie Luo, and Qixing Huang.
\newblock Starmap for category-agnostic keypoint and viewpoint estimation.
\newblock In {\em Proceedings of the European Conference on Computer Vision
  (ECCV)}, pages 318--334, 2018.

\bibitem{zhou2019context}
Yizhou Zhou, Xiaoyan Sun, Zheng-Jun Zha, and Wenjun Zeng.
\newblock Context-reinforced semantic segmentation.
\newblock In {\em Proceedings of the IEEE Conference on Computer Vision and
  Pattern Recognition}, pages 4046--4055, 2019.

\bibitem{zhu2014single}
Menglong Zhu, Konstantinos~G Derpanis, Yinfei Yang, Samarth Brahmbhatt, Mabel
  Zhang, Cody Phillips, Matthieu Lecce, and Kostas Daniilidis.
\newblock Single image 3d object detection and pose estimation for grasping.
\newblock In {\em 2014 IEEE International Conference on Robotics and Automation
  (ICRA)}, pages 3936--3943. IEEE, 2014.

\end{thebibliography}
}

\newpage
\renewcommand\thefigure{\thesection.\arabic{figure}}
\renewcommand\thetable{\thesection.\arabic{table}}
\setcounter{figure}{0}
\setcounter{table}{0}
\appendix
\section{Detailed Explanation of Pose Transformation}
\begin{figure}[htbp]
\begin{center}
\includegraphics[height=4.2cm]{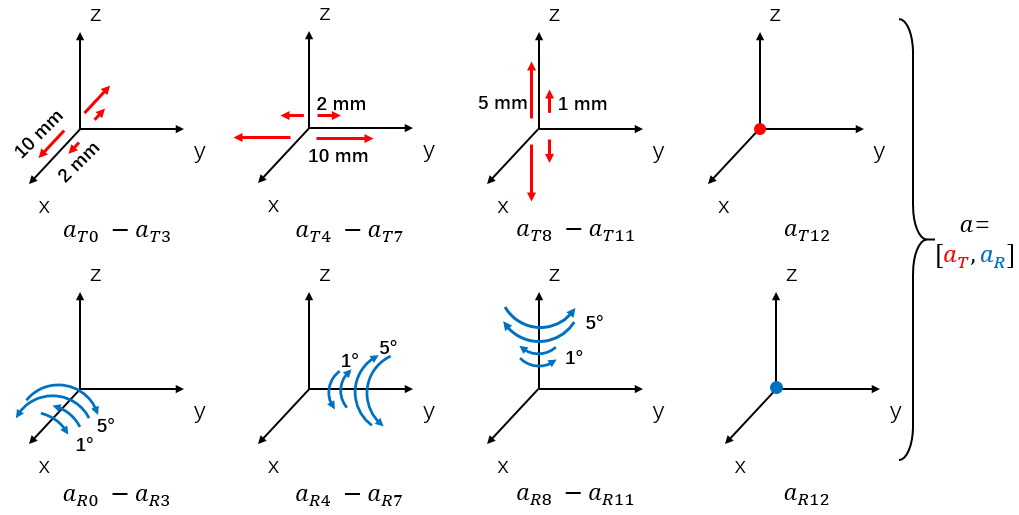}
\end{center}
   \caption{Illustration of pose transformations.}
\label{pt}
\end{figure}

As stated in the main paper (Section 3.1), the key idea is to disentangle the rotation and translation, and further discretize each individual degree of freedom for the actions. 
At each step, for each of the rotation and translation, we provide 13 actions to rotate (or translate) the object along an axis in a directional small or large step, or stay still (See the Figure~\ref{pt}).
The discretization converts the pose regression to a classification task, which tremendously reduces the training difficulty.

\section{Additional Evaluation of the Refined Poses}
Table \ref{tab:evalfreedom} shows the additional evaluation of the refined poses with AAE initialization on the whole LINEMOD test set. The average depth error is 1.66cm, a bit larger than error on axis x and y. For rotation the error on elevation and in-plane tilt is smaller than on azimuth. We observe that the in-plane rotation exhibits more significant changes in 2D masks, which may explain the higher accuracy on this DoF.
\begin{table}[htbp]
  \centering
  \resizebox{82mm}{18mm}{
    \begin{tabular}{r|r|rrr|rrr}
    \toprule
    \multicolumn{1}{r}{} &       & \multicolumn{3}{c|}{Translation} & \multicolumn{3}{c}{Rotation} \\
\cmidrule{3-8}    \multicolumn{1}{r}{} &       & \multicolumn{1}{r}{x} & \multicolumn{1}{r}{y} & \multicolumn{1}{r|}{z} & \multicolumn{1}{r}{Rx} & \multicolumn{1}{r}{Ry} & \multicolumn{1}{r}{Rz} \\
    \midrule
    \multicolumn{2}{r|}{Mean(cm, \degree)  } & 0.21  & 0.21  & 1.66  & 12.55  & 4.92  & 6.37  \\
    \multicolumn{2}{r|}{Std(cm, \degree)} & 0.31  & 0.29  & 1.95  & 30.42  & 8.57  & 18.52  \\
    \midrule
    \multirow{3}[2]{*}{\makecell[r]{Acc\\(\%)} } & 2(cm, \degree) & 99.53  & 99.67  & 72.72  & 33.06  & 41.64  & 56.66  \\
          & 5(cm, \degree) & 99.99  & 99.99  & 94.49  & 59.47  & 70.36  & 77.99  \\
          & 10(cm, \degree) & 100.00  & 100.00  & 99.25  & 77.80  & 88.35  & 88.00  \\
    \bottomrule
    \end{tabular}} %
    \caption{Accuracy of 6 degrees of freedom. Rx, Ry, Rz represent  azimuth, elevation, and in-plane rotation respectively.}
  \label{tab:evalfreedom}%
\end{table}%

\section{Results with DPOD~\cite{zakharov2019dpod} as initial pose}
Instead of AAE, we utilize PFRL to refine DPOD-syn’s~\cite{zakharov2019dpod} initial poses provided by the author to make a fair comparison. The results of recall scores on ADD are shown in Table \ref{tab:dpodinit}. When using the same initial poses, our method performs generally better than DPOD-syn.

\begin{table}[htbp]
  \centering
  \resizebox{84mm}{17mm}{
    \begin{tabular}{r|c|c|c|c|c|c|c}
    \toprule
    Metric & \multicolumn{7}{c}{Recall scores (\%) on ADD} \\
    \midrule
    Class & ape   & bv.   & cam   & can   & cat   & driller & duck \\
    \midrule
    DPOD-refine & 52.12  & 64.67  & 22.23  & 77.51  & 56.49  & 65.23  & 49.04 
  \\
    PFRL  & 69.26  & 78.68  & 27.77  & 77.16  & 64.52  & 79.90  & 48.24   \\
    \midrule
    Class & egg.  & glue  & hol.  & iron  & lamp  & phone & \textbf{Mean} \\
    \midrule
    DPOD-refine & 62.21  & 38.94  & 25.55  & 98.43  & 58.35  & 33.79  & 54.20   \\
    PFRL  & 67.68  & 37.73  & 27.87  & 88.00  & 73.67  & 37.51  & \textbf{59.85}  \\
    \bottomrule
    \end{tabular}}
    \caption{DPOD-refine/PFRL with DPOD init.}
  \label{tab:dpodinit}%
\end{table}%

\section{Generalization Ability}
To test our method's generalization ability, we evaluate the model on testing objects different from the training object on T-LESS dataset. Specifically, we trained 3 models on object 19-21 and test them on object 6-10 with AAE initialization separately. As shown in Table \ref{tab:Generalization}, the three models all improve the recall of VSD on 5 unseen objects for about 16\%-18\%, which shows our method can well generalize to unseen objects.
\begin{table}[htbp]
  \centering
  \resizebox{84mm}{21mm}{
    \begin{tabular}{r|ccc|ccc|ccc}
    \toprule
    \multicolumn{1}{l|}{Metric} & \multicolumn{9}{c}{Recall scores (\%) on VSD} \\
    \midrule
    Test Obj & \multicolumn{3}{c|}{6} & \multicolumn{3}{c|}{7} & \multicolumn{3}{c}{8} \\
    Train Obj & 19    & 20    & 21    & 19    & 20    & 21    & 19    & 20    & 21 \\
    \midrule
    AAE   & \multicolumn{3}{c|}{52.3 } & \multicolumn{3}{c|}{36.6 } & \multicolumn{3}{c}{22.1 } \\
     +PFRL & 59.1  & 62.5  & 56.9  & 51.0  & 52.1  & 51.2  & 42.1  & 42.3  & 42.7  \\
    \midrule \midrule
    Test Obj & \multicolumn{3}{c|}{9} & \multicolumn{3}{c|}{10} & \multicolumn{3}{c}{\textbf{Mean}} \\
    Train Obj & 19    & 20    & 21    & 19    & 20    & 21    & 19    & 20    & 21 \\
    \midrule
    AAE   & \multicolumn{3}{c|}{46.5 } & \multicolumn{3}{c|}{14.3 } & \multicolumn{3}{c}{34.3 } \\
     +PFRL & 54.9  & 56.9  & 54.4  & 47.0  & 50.5  & 46.9  & 50.8  & 52.9  & 50.4  \\
    \bottomrule
    \end{tabular}}%
    \caption{Recall of VSD on T-LESS objects 6-11 for the model trained on object 19-21.}
  \label{tab:Generalization}%
\end{table}%

\section{Class-Agnostic Training}
The class-specific training setting in the original manuscript was adopted considering that the RL training is quite time-consuming and difficult to converge, especially in the case of multiple objects. We conducted class-agnostic training with the same network structure on LineMOD dataset, in which we just replaced the training data from one object to all 13 objects. As shown in Table \ref{tab:cls_agn}, although the class-agnostic training result can not compare with the class-specific training, it can still bring appreciable improvement to the initial poses.
\begin{table}[htbp]
  \centering
  \resizebox{84mm}{17mm}{
    \begin{tabular}{r|c|c|c|c|c|c|c}
    \toprule
    Metric & \multicolumn{7}{c}{ADD(\%)} \\
    \midrule
    Class & ape   & bv.   & cam   & can   & cat   & driller & duck \\
    \midrule
    AAE   & 3.96  & 20.92  & 30.47  & 35.87  & 17.90  & 23.99  & 4.86  \\
    Cls Agnos & 22.00  & 56.26  & 14.02  & 53.84  & 44.21  & 43.41  & 32.68  \\
    \midrule
    Class & egg.  & glue  & hol.  & iron  & lamp  & phone & \textbf{Mean} \\
    \midrule
    AAE   & 81.01  & 45.49  & 17.60  & 32.03  & 60.47  & 33.79  & 31.41  \\
    Cls Agnos & 87.51  & 63.22  & 25.78  & 55.87  & 91.17  & 38.43  & 48.34  \\
    \bottomrule
    \end{tabular}}%
  \caption{Class agnostic training results with AAE initialization.}
  \label{tab:cls_agn}%
\end{table}%

\section{Details of Optimization Rules}
\subsection{On Policy Part}
We employ the proximal policy optimization algorithm \cite{schulman2017proximal} as the basic update rule.
Let the 6D pose estimation procedure with one RGB image have $K$ frames in total. 
At each time step $k$, let $\mathbf{s}_k$ denote the current state, then the relative SE(3) transformation $\mathbf{a}_k=[\mathbf{a}_R|\mathbf{a}_t]$ can be sampled from the network output distribution with input $\mathbf{s}_k$. 
$\pi_\theta$ denotes the current network output distribution, and $\pi_{\theta_\text{old}}$ denotes the network output distribution when $\mathbf{a}_k$ was sampled. $V(\mathbf{s}_k)=\mathbb{E}_{\mathbf{a}_k,\mathbf{s}_{k+1}...}[\sum^\infty_{l=0} \gamma^l r_{k+l}]$ is the value function, which is meant to be the expected cumulative reward under current state $\mathbf{s}_k$. $V_\theta$ denotes $V$ estimated by another network with input $\mathbf{s}_k$ that shares the same weights as the action network except for the last layer. 
The clipped surrogate objective can be written as:
\begin{equation}\label{cso}
L_{c}(\theta)=\hat{\mathbb{E}}[\min(\xi_k(\theta)\hat{A}_k,\text{clip}(\xi_k(\theta),1-\epsilon, 1+\epsilon)\hat{A}_k)],
\end{equation}
where
\begin{equation}\label{ratio}
\xi_k(\theta)=\frac{\pi_\theta(\mathbf{a}_k|\mathbf{s}_k)}{\pi_{\theta_{\text{old}}}(\mathbf{a}_k|\mathbf{s}_k)}.
\end{equation}
The $\hat{A}_k$ in Eq.(\ref{cso}) is the advantage estimator defined as: $\hat{A}_k=-V(\mathbf{s}_k)+ r_k+\gamma r_{k+1}+...+ \gamma^{K-k+1}r_{K-1}+\gamma^{K-k}V(\mathbf{s}_K)$. The value loss can be written as:
\begin{equation}
L_v(\theta) = (V_\theta(\mathbf{s}_k)-V_{\text{targ}})^2,
\end{equation}
where
\begin{equation}
V_{\text{targ}}=r_k+\gamma r_{k+1}+...+\gamma^{K-k}V_{\theta_{\text{old}}}(\mathbf{s}_K).
\end{equation}
And the entropy regularization term to encourage adequate exploration can be written as:
\begin{equation}
L_e(\theta)=\pi_\theta\log{\pi_\theta}.
\end{equation}
The on-policy update loss $L_{\text{on}}$ can be written as:
\begin{equation}
L_{\text{on}} = L_c+\lambda_vL_v+\lambda_eL_e.
\end{equation}
\subsection{Off Policy Part}
We introduce the V-trace target from \cite{espeholt2018impala} to use samples more efficiently with an off-policy update and give value function a more accurate estimation. 
The n-step V-trace target can be written as:
\begin{equation}
V_{\text{trace}} = V(\mathbf{s}_k) +\sum^{k+n-1}_{q=k}\gamma^{q-k}(\prod^{q-1}_{i=k}c_i)\delta_qV,
\end{equation}
where
\begin{equation}\label{vtrace2}
\begin{aligned}\delta_q V=\rho_q(r_q+\gamma V(\mathbf{s}_{q+1})-V(\mathbf{s}_q)),\\
\rho_q=\min(\overline{\rho},\frac{\pi_\theta(\mathbf{a}_q|\mathbf{s}_q)}{\pi_{\theta_\text{old}}(\mathbf{a}_q|\mathbf{s}_q)}),\\c_i=\min(\overline{c},\frac{\pi_\theta(\mathbf{a}_i|\mathbf{s}_i)}{\pi_{\theta_\text{old}}(\mathbf{a}_i|\mathbf{s}_i)}).\end{aligned}
\end{equation}
In Eq.(\ref{vtrace2}), $\rho_q$ and $c_i$ are truncated importance sampling weights, and the truncated parameters $\overline{\rho}\ge \overline{c}$. 
The off-policy value function loss is:
\begin{equation}
L_{\text{off}}(\theta)=(V_\theta(\mathbf{s}_k)-V_{\text{trace}})^2.
\end{equation}

\section{Results on T-LESS}
\begin{figure*}[!htbp]
\begin{center}
\includegraphics[height=16cm]{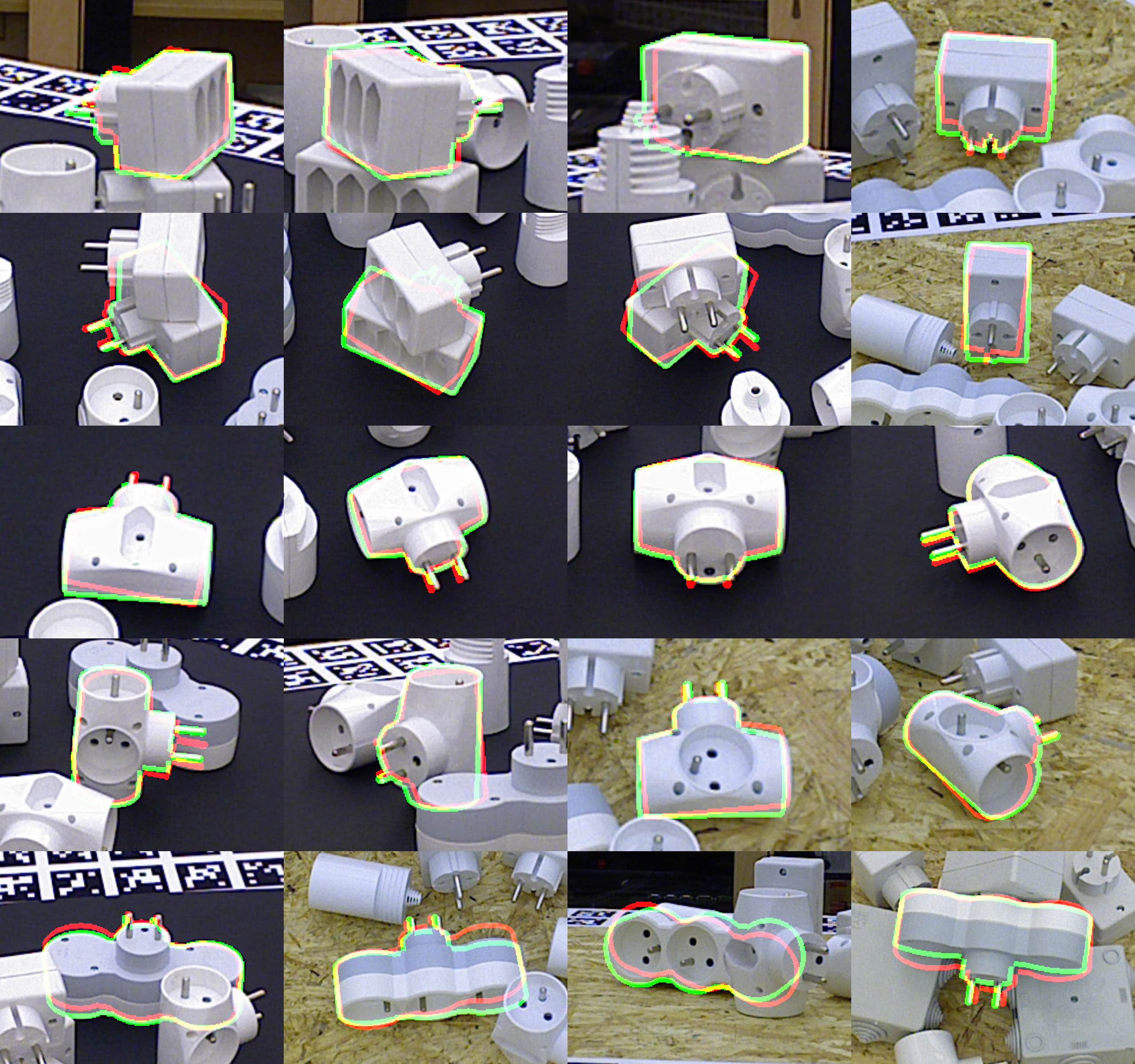}
\end{center}
   \caption{Qualitative results of AAE initial poses~\cite{sundermeyer2018implicit} and our refined poses on the T-LESS dataset. 
   The red and green lines represent the contours of the initial poses and our refined poses respectively.}
\label{tlessm}
\end{figure*}
Fig. \ref{tlessm} shows some qualitative results of objects 19-23 on T-LESS dataset~\cite{hodan2017t}.

\end{document}